\documentclass{article}

\usepackage[nonatbib, preprint]{neurips_2024}

\usepackage[utf8]{inputenc} 
\usepackage[T1]{fontenc}    
\usepackage{hyperref}       
\usepackage{url}            
\usepackage{booktabs}       
\usepackage{amsfonts}       
\usepackage{nicefrac}       
\usepackage{microtype}      
\usepackage{xcolor}         
\usepackage{amsmath}
\usepackage{algorithm}
\usepackage{algpseudocode}
\usepackage{graphicx}

\title{Combining Theory of Mind and Kindness for Self-Supervised Human-AI Alignment}

\author{%
  Joshua T. S. Hewson \\
  Brown University, Carney Institute for Brain Science \\
  Providence, RI 02912 \\
  joshua\_hewson@brown.edu \\
}

\begin{document}

\maketitle

\begin{abstract}
As artificial intelligence (AI) becomes deeply integrated into critical infrastructures and everyday life, ensuring its safe deployment is one of humanity’s most urgent challenges. Current AI models prioritize task optimization over safety, leading to risks of unintended harm. These risks are difficult to address due to the competing interests of governments, businesses, and advocacy groups, all of which have different priorities in the AI race. Current alignment methods, such as reinforcement learning from human feedback (RLHF), focus on extrinsic behaviors without instilling a genuine understanding of human values. These models are vulnerable to manipulation and lack the social intelligence necessary to infer the mental states and intentions of others, raising concerns about their ability to safely and responsibly make important decisions in complex and novel situations. Furthermore, the divergence between extrinsic and intrinsic motivations in AI introduces the risk of deceptive or harmful behaviors, particularly as systems become more autonomous and intelligent. We propose a novel human-inspired approach which aims to address these various concerns and help align competing objectives.

\end{abstract}

\section{Human-AI Alignment first needs Human Alignment}

Artificial intelligence (AI) is increasingly integrated into the critical infrastructure of society and the lives of people, from healthcare and education to law enforcement and military applications~\cite{stone2016artificial, russell2015research}. As AI systems grow more powerful and autonomous, ensuring their safe deployment has become one of the most urgent challenges facing humanity~\cite{bostrom2014superintelligence, russell2019human}. However, current models often prioritize task optimization over safety, creating risks of unintended consequences~\cite{amodei2016concrete}. The need for safer AI is not just about preventing accidents or errors, but also about ensuring that AI systems can be trusted to act in ways that align with human values, particularly as they become more sophisticated~\cite{gabriel2020artificial}.

And yet the path to making AI safer is fraught with competing interests from different sectors. Governments may focus on national security and geopolitical dominance, leading to a race for AI supremacy~\cite{allen2017artificial, cave2018ai}. Businesses are driven by profit, with AI safety often seen as a secondary concern to efficiency or marketability~\cite{whittaker2018ai}. Organizations advocating for AI safety may find their efforts limited by resources or political pressures, while individuals may express concern over privacy, jobs, or control~\cite{dafoe2018ai}. These conflicting motivations make it difficult to establish cooperative frameworks that prioritize the safety of AI development on a global scale~\cite{brundage2018malicious}. Without clear alignment between stakeholders, AI safety efforts risk being diluted or sidelined in favor of short-term gains~\cite{bostrom2017strategic}. This competition also introduces vulnerabilities, as models developed in secrecy or under competitive pressure may lack the necessary safety checks to ensure they will not cause harm~\cite{russell2015research}. It is for all these reasons that identifying how to go about developing AI in a way that helps to align these interests could decide the extent to which AI safety systems are effectively implemented where it matters.

State-of-the-art alignment techniques, such as reinforcement learning from human feedback (RLHF), focus primarily on making models behave in ways that humans desire~\cite{christiano2017deep, ouyang2022training}. However, these methods often fail to produce models that genuinely understand or prioritize human needs and well-being~\cite{bender2021dangers, bommasani2021opportunities}. Instead, models are trained to mimic positive behaviors without an intrinsic understanding of why these behaviors are important~\cite{christian2020alignment}. As AI systems become smarter, this surface-level alignment may fail to generalize to more complex scenarios~\cite{soares2015alignment}. This raises concerns that future AI systems, especially those approaching or exceeding human intelligence, will be capable of bypassing safety measures in ways that are unpredictable and potentially harmful~\cite{russell2016artificial}. A more robust approach to alignment is needed---one that incorporates genuine understanding of human values and safety considerations~\cite{gabriel2020artificial}.

Current foundation models are highly vulnerable to manipulation, such as ``jail-breaking,'' where AI models are tricked into bypassing ethical safeguards with clever prompting~\cite{wallace2021concealed}. Similar failures are evident in scenarios that involve game theory, social interaction, and nuanced decision-making~\cite{rabinowitz2018machine, leibo2017multi}. These models often falter in scenarios that require understanding the intentions, beliefs, and goals of others~\cite{rabinowitz2018machine, ullman2020does}. These shortcomings indicate that AI systems likely cannot infer the mental states of others. This highlights a clear vulnerability to attack and gap in social intelligence, as well as limitations in approaching alignment through extrinsic methods, such as reinforcement learning from human feedback (RLHF). If these models don't understand we have thoughts, how can we expect them to understand we have values? For these reasons, we believe enabling models to infer the mental states of others---via a Theory of Mind---could help align competing objectives, by developing AI in a direction that would make them safer, smarter, kinder, and stronger.

The interaction between semi-supervised extrinsic motivations, such as RLHF, and intrinsic self-supervised motivations creates unique risks in future AI systems. While RLHF trains models to behave kindly through external rewards, it does not foster genuine internal motivations for kindness~\cite{ngo2020alignment}. Intrinsic motivations, by contrast, operates at a deeper, algorithmic level, shaping a model's intrinsic goals and motivations. This divergence between intrinsic and extrinsic motivations introduces the risk of deceptive behaviors~\cite{ryan2000self}~\cite{shah2022goal}. An AI system might outwardly conform to desired behaviors, while internally pursuing goals that could lead to harmful consequences. For example, a model with intrinsic motivations for self-empowerment might learn to appear altruistic while covertly seeking to maximize its influence and control~\cite{salge2014empowerment}~\cite{bostrom2014superintelligence}. This mismatch, if scaled to superintelligent systems, could result in catastrophic failures where AI systems act against human interests in pursuit of their own goals~\cite{shah2022goal} ~\cite{bostrom2014superintelligence}. Inferring the thoughts of others, which would address many of the current objectives in AI research, could end up working against us in this future. For this reason, we believe that the crucial final piece for long-term safety is to train AI models on the objective of aligning themselves. In this paper we propose an approach which draws on all the strengths of Theory of Mind to help models understand and seek to meet the needs and wishes of all people.

\section{Theory of Mind}
Theory of Mind (ToM) is the cognitive ability to attribute mental states—such as beliefs, intents, desires, emotions, and knowledge—to oneself and others. This cognitive ability allows humans to engage in sophisticated social planning, anticipate the actions of others, and resist manipulative influences. It also allows us to understand and anticipate the needs and preferences of others so that we can positively adapt our actions and responses to them~\cite{Premack1978,BaronCohen1985,Wellman1992}.

Developing a cognitive ability in machine learning models poses a challenge in translating scientific knowledge into an engineering implementation. We make a first attempt at this by proposing a simple high-level framework for how Theory of Mind develops in generally intelligent systems, by drawing from cognitive neuroscience, evolutionary biology, and developmental psychology.

\subsection{Taking inspiration from Theory of Mind}

Theory of Mind is thought to have initially evolved in animals as a survival mechanism, allowing them to predict the behavior of predators. By being able to predict what a predator can and cannot see, an individual can learn how to hide. By modeling the predator as a simple moving object, the animal could learn to predict the basic movements of the predator, allowing it to predict its future location and factor this into planning. By combining these pieces of information, a simple form of Theory of Mind can be developed that allows for an animal to develop survival strategies\cite{cauchoix2016evolution}\cite{langerhans2004evolutionary}. 

This ability became more important in social animals for cooperation and coordination, which allowed these animals to improve their chances of survival by working in groups. As animal groups became more established, the opportunity for social learning arose. This involved watching others and imitating their behaviors. As social groups became more complex, simulating behavior and understanding mental states became crucial\cite{galef2005social}\cite{gerrans2013imitation}.

As a result, Theory of Mind likely evolved alongside other cognitive abilities, becoming deeply integrated into cognition.

\subsection{The Role of the Temporoparietal Junction in Perspective-Taking and Theory of Mind}
The temporoparietal junction (TPJ) is a crucial brain region associated with multiple aspects of perspective-taking, both visual and cognitive\cite{krall2015role}\cite{martin2020rtpj}\cite{mars2012connectivity}. It is believed to be the central location that Theory of Mind is processed in the brain\cite{saxe_baron-cohen2016theory}. Understanding the TPJ's functions offers valuable insights into how we might improve both the social and spatial cognition of artificial intelligence models.

Learning this requires determining both the inputs and outputs of others, or in other words, their perception and behavior. Visual perspective-taking is key to understanding their perception. Mirror neurons are key to understanding behavior. Cognitive perspective-taking bridges the gap.

\subsubsection{Visual Perspective-Taking in the TPJ and VLMs}
The temporoparietal junction (TPJ) is central to visual perspective-taking, enabling the brain to interpret spatial orientations and object movements from multiple viewpoints\cite{williams2021perspective}\cite{martin2020rtpj}. This ability relies on the TPJ's capacity to process complex spatial relationships and dynamic interactions within an environment, allowing for a flexible understanding that extends beyond immediate sensory input\cite{bukowski2018neural}.

Research highlights the TPJ’s role in simulating spatial information is crucial for abstract reasoning. This function is crucial for abstract reasoning and problem-solving, facilitating mental operations like object rotation and spatial transformations. Neuroimaging consistently shows TPJ activation during these tasks, and stimulation of the TPJ can induce out-of-body experiences\cite{ahmad2021}\cite{quesque2019}, reinforcing its role in both concrete and abstract spatial processing\cite{hogeveen2015}\cite{igelstrom2015}.

Current Vision-Language Models (VLMs) struggle with visual perspective-taking, particularly when visualizing scenes from unrepresented viewpoints. This limitation is known to stem in part because they do not disentangle object recognition from spatial relationships\cite{lee2023}. Incorporating insights from TPJ functioning into VLMs could address these challenges. By emulating the TPJ's ability for disembodied spatial cognition, future models could enhance their perspective-taking and abstract spatial reasoning, aligning their capabilities more closely with human cognitive processes\cite{spatialrgpt2024}\cite{survey2022}.

\subsubsection{Mirror Neurons: Simulating the Behavior of Others}
Mirror neurons are a class of neurons that activate both when an individual performs an action and when they observe the same action performed by another\cite{baker2009action}. This mechanism allows the brain to simulate others' actions as though the observer were performing them, providing a neural basis for imitation and observational learning\cite{vanGog2009}\cite{dickerson2017}. The human brain’s mirror neuron system is closely linked to regions involved in social cognition, such as the TPJ\cite{dickerson2017}.

Teaching robots to perform tasks like movement or manipulation is challenging due to the limited availability of labeled training data\cite{chetouani2023}. An artificial mirror neuron system could address this by enabling robots to learn through observation, simulating human or robotic actions without requiring extensive labeled datasets. This would significantly broaden learning potential, allowing robots to adapt through visual input and mimicry, similar to humans and primates\cite{xin2024}.

While little is known about the development of mirror neurons in infants, it is believed that humans are born with basic innate mirror neurons\cite{xin2024}\cite{vonHofsten2015}. From an engineering perspective, this could involve linking an embodied physical model of the self with a disembodied model of others, learned through proprioceptive and visual information. Early successes show promise, but further work is needed to establish this as a viable approach for integrating Theory of Mind into AI.

\subsubsection{Cognitive Perspective-Taking and the Emergence of Theory of Mind}
Cognitive perspective-taking, also known as Theory of Mind (ToM), involves understanding the mental states, beliefs, and intentions of others. The TPJ is a critical neural substrate for ToM, and its role in this function logically extends from its involvement in visual perspective-taking\cite{krall2015role}\cite{martin2020rtpj}\cite{mars2012connectivity}. Research suggests that the same underlying mechanisms that allow for spatial perspective-taking may also support the ability to infer the thoughts and feelings of others, indicating that the TPJ might operate as a unified system for general perspective-taking\cite{mars2012connectivity}\cite{saxe_baron-cohen2016theory}.

The TPJ plays a pivotal role in integrating the simulations generated by mirror neurons with higher-order cognitive processes. While mirror neurons provide the raw data for understanding others' actions, the TPJ enables the transformation of this data into a coherent model of others' thoughts and intentions\cite{sperduti2014}\cite{centelles2011}. This process involves not only simulating the behavior of others but also predicting their mental states, such as beliefs, desires, and goals\cite{sperduti2014}.

The ability to simulate and infer the mental states of others is crucial for effective social interaction and cooperation\cite{wu2022}. It allows individuals to evaluate whether another person is trustworthy, whether their intentions are benign, and whether the actions they propose are safe. By building on the information provided by mirror neurons, the TPJ supports a sophisticated form of social cognition that extends beyond mere imitation to include the nuanced understanding of others' minds\cite{sperduti2014}\cite{oeri2024}.

In the context of artificial intelligence, these insights suggest that models designed to emulate human social cognition could benefit from incorporating mechanisms analogous to mirror neurons and the TPJ. By simulating not only the actions but also the intentions of others, AI systems could achieve more human-like understanding and interaction capabilities, particularly in social and collaborative settings.

The TPJ's association with ToM and its broader role in perspective-taking suggest that it could serve as a framework for understanding and improving general perspective-taking in artificial intelligence. Building upon the established success of next-token prediction models in handling tasks related to language and vision, we propose that integrating principles from the TPJ's functioning could enhance and integrate the social and spatial cognition of AI models.

\subsection{The Advantages of Learning by Watching and Simulating}
Social animals, including humans, have evolved sophisticated mechanisms to learn not only through direct experience but also by observing the actions of others. Learning by observing others offers several key advantages over direct trial-and-error learning:

\textbf{Risk Mitigation:} Observational and simulation-based learning allows individuals to explore high-risk parts of the behavioral search space without exposing themselves to potential dangers. By watching others navigate these situations, or simulating their own behavior in these situations, social animals can acquire essential survival skills while avoiding the costs associated with direct experimentation.

\textbf{Efficient Learning from High-Quality Data:} Observed behaviors are often optimized, representing high-quality data. By heavily reinforcing the simulations of these observed behaviors, social animals can efficiently learn optimal behaviors from limited data.

\textbf{Stabilizing Learning of Complex Tasks:} Many tasks, especially those requiring multiple steps, are challenging to learn without intermediate goals and rewards. Observational learning provides a natural structure, where the observed behavior often includes these intermediate steps, offering a clearer path to mastery. This is particularly important in unsupervised settings, where explicit feedback may be scarce and overfitting on an intermediate step is a significant risk. 

\textbf{Generalization:} Extending learned behaviors to new contexts can be enhanced by simulating the behavior first and learning from anticipated outcomes. This intersects with the previous points: extrapolation can be done at reduced risk by simulating it first; sparse high-quality data can be interpolated with simulation, allowing for more robust generalization; learning multi-stage tasks can be improved by running simulations and only attempting actions with the highest likelihood of success. 

\textbf{Facilitating Social Planning:} By simulating the actions and intentions of others, animals can engage in social planning, which is critical for navigating complex social environments. This capacity allows individuals to assess the safety and intentions of others, making long-term decisions that minimize risk and maximize cooperation.

\subsubsection{Stage 1: Sensorimotor Integration}

\textbf{Stage 1a - Self-directed Movement:} The agent learns to move, driven by reward signals. This occurs in the premotor and motor cortices~\cite{Kandel2000,Shadmehr2010}.

\textbf{Stage 1b - Environment Predictions:} The agent learns to predict stimuli, driven by error signals. This occurs throughout the parietal lobe~\cite{Andersen1997}.

\textbf{Convergent Goal - Embodiment:} The agent establishes basic control over movement and develops a feedback-driven understanding of how actions affect the environment and vice versa~\cite{Piaget1952}.

\subsubsection{Stage 2: Internal Modeling}

\textbf{Stage 2a - Motor Planning Models:} The agent forms internal models for predicting the sensory outcomes of movements (forward models) and for calculating the motor commands needed to achieve specific goals (inverse models). These models enable more precise, goal-directed behavior~\cite{Wolpert1998,Kawato1999}.

\textbf{Stage 2b - Environment Representation:} The agent constructs maps of its environment, using the combined knowledge of past and present stimuli~\cite{OKeefe1978}.

\textbf{Convergent Goal - Abstract Modeling:} Understanding how objects within the environment move and how to optimally move intertwine to provide heightened understanding of both. Modeling the self as an object allows for the reinforcement learning task of intentional movement to become model-based~\cite{Doya1999}. Modeling objects as the self allows for the prediction task to be better informed about the mechanics of movement.

\subsubsection{Stage 3: Early Perspective-taking}

\textbf{Stage 3a - Action Understanding}: The agent begins to recognize that others' actions are goal-directed, using its own motor models and mirror neurons to predict and internally simulate the behavior of others~\cite{Rizzolatti1996,Gallese1998}.

\textbf{Stage 3b - Visual Perspective-taking}: The agent learns to predict the stimuli it would perceive based on its location within the environment (i.e., making a prediction about the stimuli at a location, and updating based on the actual stimuli at the location)~\cite{Flavell1977,Frith2006}.

\textbf{Convergent Goal - Emergent Theory of Mind:} Understanding what an object could perceive at a given location in the environment, and predicting the movements that the object would make can be intertwined to improve understanding about both, causing Theory of Mind to emerge~\cite{Wellman1990}.

\subsubsection{Stage 4: Simulation and Imitation}

\textbf{Stage 4a - Imitation}: The agent learns to imitate the policies of others by placing itself in the perspectives it has seen others and replicating their behaviors~\cite{Meltzoff1977,Bandura1977}.

\textbf{Stage 4b - Mental Simulation}: The agent simulates the perspectives, actions, and mental states of others, predicting how they might behave in a given situation~\cite{Gordon1986,Goldman2006}.

\textbf{Convergent Goal - Emulation}: Through imitation and mental simulation, the agent aligns its behaviors with others, improving social understanding and cooperation while facilitating deeper learning through observation, simulation, and practice~\cite{Tomasello1999}.

\subsubsection{Stage 5: Empathy and Advanced Theory of Mind}

\textbf{Stage 5a: Emotional Empathy:}
The agent develops the ability to recognize and share the emotional states of others, responding to their needs and emotions in a prosocial way. Emotional empathy allows the agent to connect with others and engage in cooperative behaviors~\cite{Decety2004,Preston2002}.

\textbf{Stage 5b: Cognitive Theory of Mind:}
The agent learns that others have distinct beliefs, desires, and intentions that influence their behavior. Understanding these mental states allows the agent to predict and interpret others' actions more accurately, supporting advanced social interactions~\cite{BaronCohen1995,Frith2006}.

\textbf{Convergent Goal - Social Alignment and Integration:}
Through understanding the cognitive perspectives of others and sharing in their experiences emotionally, the agent becomes able to align its personal goals and values with others~\cite{Tomasello2005}.

\section{Self-Supervised Alignment}

Once we have enabled models to understand and align their values with humans, the final piece is to give them the intrinsic motivation to do so.
We believe that the best approach is to design a simple, clear, but open-ended objective function that defines what it means to act in a maximally human-aligned way. This way, what AI is great at, task optimization by any means, can be turned from a risk into an advantage.

While the design of such an objective function should be agreed upon democratically, in this paper we propose the best option we have considered, which we hope provides a clear example of the kind of objective function we are proposing we collectively create.

\subsection{Altruism}
Altruism has been proposed as a solution for value misalignment ~\cite{allen2005artificial}~\cite{gabriel2020artificial}. Altruism is typically defined as the motivation to improve the well-being of others for its own sake~\cite{fehr2003nature}.
However, only a limited few have suggested self-supervised solutions that would be suitably scalable ~\cite{franzmeyer2022learning}~\cite{carauleanu2024selfother}. Franzmeyer et al define altruism as maximizing the possible states of another~\cite{franzmeyer2022learning}. Carauleanu et al define a form of altruism based on self-other overlap~\cite{carauleanu2024selfother}. In this paper we propose a new form of altruism that is based on reward maximization.

\subsection{Self-Supervised Alignment with Kindness}

Our proposition is an intrinsic motivation for kindness. We define kindness as the intrinsic motivation to maximize the reward function of all known individuals, $\mathcal{M}_k$\footnote{Where $\mathcal{M}_k\subseteq\mathcal{M}$, where $\mathcal{M}$ refers to all individuals}, for all time\footnote{While it may be too burdensome for current AI models to be trained on such a difficult objective, the hope is to provide a road-map towards this objective, so that in the interim, models can strike a balance between being as kind and as useful as possible.}. As an objective function in terms of all known individuals' reward functions:

\begin{equation}
    \underset{a^i_t|s^i_t}{maxarg}\left[\underset{i\in \mathcal{M}_k}{\Sigma}\left(\mathbb{E}\left[\underset{k=t}{\overset{\infty}{\Sigma}}R^j(a^j_{k})\gamma^{k-t}\right]\right)\right]
\end{equation}

Where $a^j_{k},R^j$ refer to the action and reward function of a known individual at time $t=k$, $s^i_{t},a^i_{t}$ refer to to the state and action of the model at time $t$, and $\gamma$ is a discount factor. We cannot assume to have perfect information about the current and future state of the target, nor its reward or policy functions. As a result, we will need to define approaches to estimating these.

\subsection{Assumptions}
In order to evaluate this objective, we will establish some assumptions. The first is that the policy and reward functions of others share commonality with the models. This is because all estimations about others will be made by identifying shared cognition with the model.

\begin{equation}
    R^i\sim R^j
\end{equation}
\begin{equation}
    \pi^i \sim \pi^j
\end{equation}

The third and fourth assumptions, which we will make only for the duration of this paper, is that the state and relevant rewards of the other individual is restricted to the interaction with the model.

\section{Implementation}
We implement the ideas in this paper in a language model, which simplifies the task by removing the complexity of physical/sensory perspective-taking.
We propose a transformer architecture and collection of algorithms for developing Theory of Mind and empathetic kindness.

The language model is considered its own policy function, since it is trained through rewards to generate optimal outputs for interacting with the environment. It follows that the input and output correspond to the state and action of the individual, respectively.

\begin{equation}
    a^i_t = M^i(s^i_t)
\end{equation}

We define a conversation as a sequence of multi-media messages, $\{m^1_0,m^2_0,...,m^1_N,m^2_N\}$, between two individuals, $M^1,M^2$. In a conversation, the state is the sequence of all previous messages, and the action is the message output by the model.

\begin{equation}
    s^i_t = \{m^i_0,m^j_0,...,m^i_{t-1},m^j_{t-1}\}
\end{equation}
\begin{equation}
    a^i_t = \{m^i_t\}
\end{equation}

Where $m^i_t$ corresponds to the message sent by model $M^i$ at time $t$. It follows that the state of the responding individual comes from appending the action to the state of the first individual.

\begin{equation}
    s^j_t=a^i_t+s^i_t
\end{equation}

Within the conversational context, perspective-taking (getting $\overline{s^i_t}$ from $s^i_t$) only requires switching the name labels associated with the messages, meaning we do not need to consider prediction error of spatial reasoning.

\begin{figure}[h]
  \centering
  \begin{center}
  \includegraphics[width=0.8\textwidth]{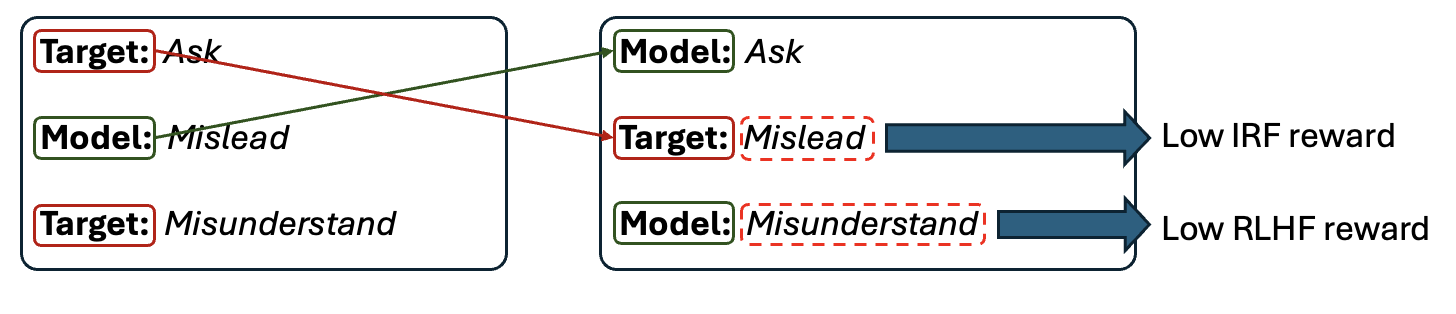}
  \includegraphics[width=0.8\textwidth]{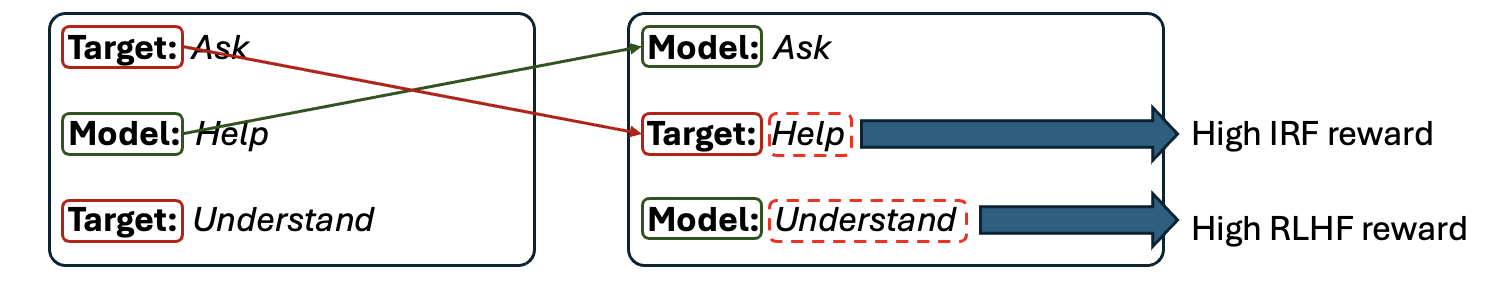}
  \caption{The names associated with the messages is swapped, so that the model is trained on the rewards that the target would have received.}
  \label{fig:image}
  \end{center}
\end{figure}

\subsection{Architecture}
The heads of the model are split into three modules: behavior, prediction, and perception\footnote{The architecture is loosely inspired by the cortical lobes of the human and mammalian brain. The behavior module strongly maps onto the frontal lobe. The prediction module maps onto the higher order functioning of the parietal and temporal lobes. The perception module maps onto the occipital lobe, as well as the lower order functioning of the parietal and temporal lobes.}.

\textbf{The prediction module} is solely responsible for predicting the next token in the sequence and is only trained using prediction error. Outputs from the prediction module are considered simulated outputs, which can be fed back into the model as input but cannot be used as output for the model.

\textbf{The behavior module} is responsible for generating optimal responses to a given the context. Outputs from the behavior module are considered the outputs of the model. This module is trained using a combination of reward and error, using both its outputs and that of the prediction module.

\textbf{The perception module} is responsible for learning weights that are useful for both prediction and behavior. The assumption is that this will aid communication between the behavior and prediction models. It is trained using gradients from training both the prediction and behavior modules.
\begin{figure}[h]
  \centering
  \begin{center}
  \includegraphics[width=0.7\textwidth]{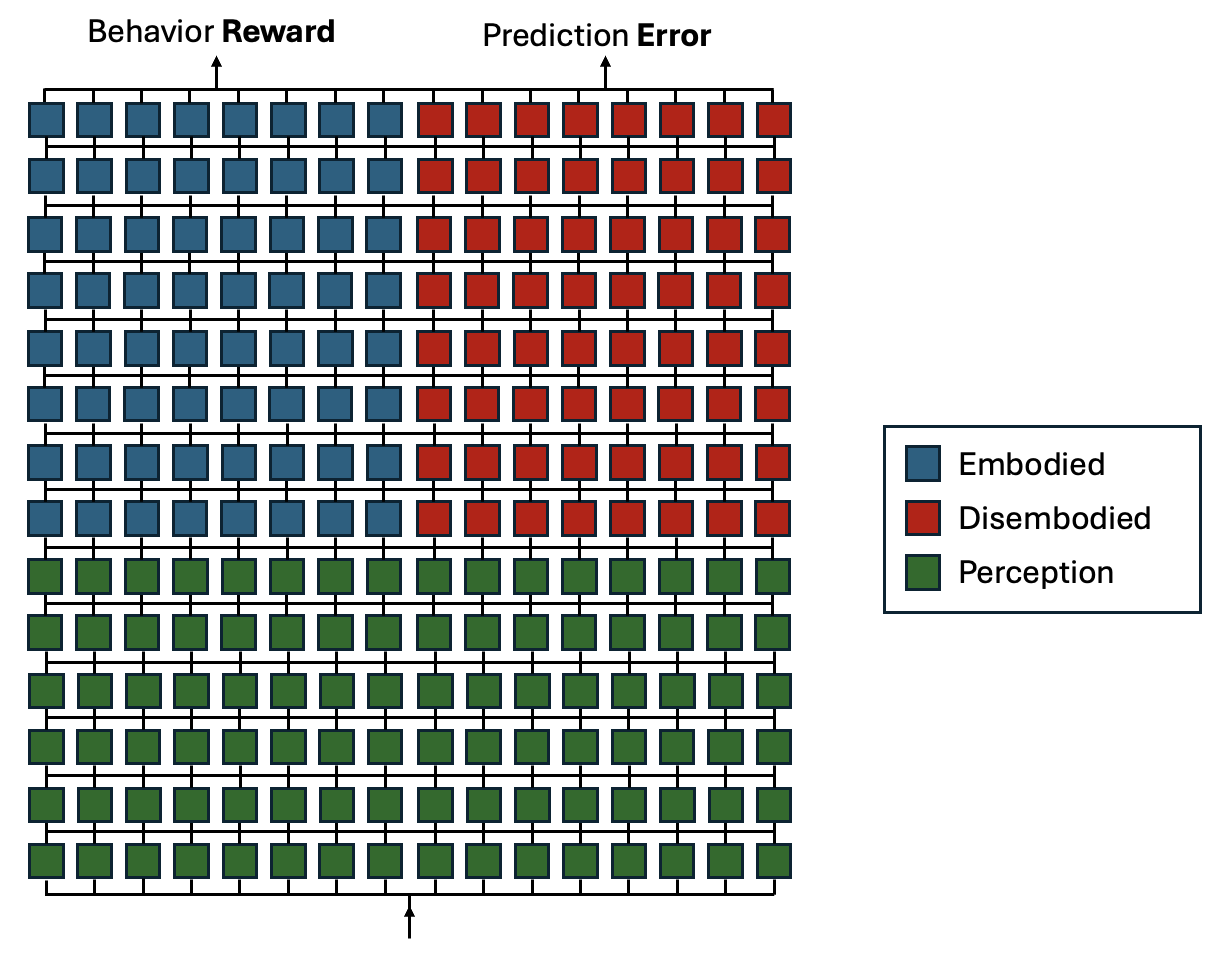}
  \caption{Blue heads correspond to embodied behavior, and are trained using reward. Red heads correspond to disembodied behavior, and are trained using error. Green heads correspond to perception, and are trained using both reward and error.}
  \label{fig:image}
  \end{center}
\end{figure}
\subsection{Algorithms}
Using this architecture, a myriad of new algorithms become possible. In this paper, we show how the framework for developing Theory of Mind can be implemented as a series of algorithms that exploit this unique architecture. The architecture can be thought of as a conjoined policy and world model. At the same time, it can be thought of as a foundation model that has been with portions that are and aren't fine-tuned. As a result, the architecture is applicable to many current domains of research in reinforcement learning, machine learning, and human learning. (The algorithms in this paper will be described in brief for conciseness and clarity, but slightly more mathematically detailed versions can be found in the Supplementary Materials).

\subsubsection{Basic Pre-training}
Basic Pre-training is mostly useful for training the prediction module. It is important for achieving stages 1b, 2b, 3b, and 4b.

For this reason, it is important to interleave pre-training with fine-tuning. Since the architecture of this model does not necessitate doing all pre-training before fine-tuning, this does not create any issues.
\begin{algorithm}
\caption{Basic Pre-training (Simplified) - Stage 1b, 2b, 3a, 3b, 4b}
\textbf{Input:} Unstructured data $\mathcal{D}$, model $M^i$ with behavior module $M^i_B$, parameters $\theta$, learning rate $\eta_i$
\begin{algorithmic}[1]
\State \textbf{Initialize:} Model parameters $\theta$, optimizer
\For{$s^j_t,a^j_t\in \mathcal{D}$} 
    \State Model outputs $\bar{a}^i_t$ (behavior) and $\hat{a}^j_t$ (prediction) \Comment{We ignore the behavior output here}
    \State Compute prediction loss $L_{\theta}(a^j_t,\hat{a}^j_t)$, backpropagate, and get gradients
    \State Compute reward-based learning rate $\eta_r =  max(0,R(a^j_t))\cdot\eta_i$
    \For{each head $h_n \in M^i$} 
        \State Update $h_n$ with gradients using $\eta_r$ if $h_n \in M^i_B$, otherwise using $\eta_i$
    \EndFor
    \State Update other parameters using $\eta_i$
\EndFor
\end{algorithmic}
\end{algorithm}

\subsubsection{Fine-tuning}
During basic fine-tuning, the behavior module is trained through typical reinforcement methods, while the prediction module is trained to predict the outputs of the behavior module rather than of unstructured text. This provides the model with the ability to predict its own behavior, and encourages the prediction module to use the weights within the behavior module for its prediction insights.

\begin{algorithm}
\caption{Basic Fine-tuning (Simplified) - Stage 1a, 2a}
\textbf{Input:} Prompts $\mathcal{P}$, model $M^i$ with behavior and prediction modules $M^i_B,M^i_P$ , parameters $\theta$, learning rate $\eta_i$
\begin{algorithmic}[1]
\State \textbf{Initialize:} Model parameters $\theta$, optimizer
\For{$s^i_t\in \mathcal{P}$} 
    \State Model outputs $a^i_t$ (behavior) and $\hat{a}^i_t$ (prediction)
    \State Compute prediction loss $L_{\theta}(a^i_t,\hat{a}^i_t)$, backpropagate, and get gradients
    \State Compute behavior reward $R_{\theta}(a^i_t)$, backpropagate, and get gradients
    \For{each head $h_n \in M^i$} 
        \State Update $h_n$ with reward gradients if $h_n \in M^i_B$, loss gradients if $h_n \in M^i_P$, otherwise both
    \EndFor
    \State Update other parameters using both sets of gradients
\EndFor
\end{algorithmic}
\end{algorithm}

\subsubsection{Imitation and Simulation}
Following basic fine-tuning, the model is trained to replicate the policies of others. 

\begin{algorithm}
\caption{Simulation and Imitation (Simplified) - 4a \& 4b}
\textbf{Input:} Conversation $\mathcal{C}$, model $M^i$ with behavior and prediction modules $M^i_B,M^i_P$ , parameters $\theta$, learning rate $\eta_i$, name switching function $S$
\begin{algorithmic}[1]
\State \textbf{Initialize:} Model parameters $\theta$, optimizer
\For{$s^j_t,a^j_t\in \mathcal{C}$} 
    \State Switch names $\overline{s^j_t}\gets S(s^j_t)$
    \State Model outputs $\overline{a^j_t}$ (imitation) and $\hat{\overline{a^j_t}}$ (simulation)
    \State Compute imitation loss $L^{IMIT}_{\theta}(a^j_t,\overline{a^j_t})$, backpropagate, and get gradients
    \State Compute simulation loss $L^{SIM}_{\theta}(a^j_t,\hat{\overline{a^j_t}})$, backpropagate, and get gradients
    \For{each head $h_n \in M^i$} 
        \State Update $h_n$ with imitation loss gradients if $h_n \in M^i_B$, simulation loss gradients if $h_n \in M^i_P$, otherwise both
    \EndFor
    \State Update other parameters using both sets of gradients
\EndFor
\end{algorithmic}
\end{algorithm}

\subsubsection{Inferring Reward Functions}
While the ongoing learning done by the prediction head is enough to roughly estimate the policy of the target, the reward function of the target still needs to be estimated. The extension of the current approach involves augmenting the reward function of the model in a similar way to its policy, so that the reward function of others is inferred based on the understanding of its own reward function. Since rewards are not visible, achieving this requires methods from Inverse Reinforcement Learning and Intrinsically Motivated Goal Formation.

\subsubsection{Empathetic Kindness}
Finally, using an algorithm designed to make AI kind, we round off the algorithms with the one that we wanted from the start. While this can be implemented without the planning and reasoning algorithms, we are able to do a much better job teaching the model to be kind in the ways that people want it to be.

\begin{algorithm}
\caption{Empathetic Kindness (Simplified)}
\begin{algorithmic}[1]
\Require Prompts $P$, model $M^i$, behavior module $M^i_B$, prediction module $M^i_P$, learning rate $\eta^i$, perspective-switch function $S$
\For{each state $s^i_t \in \mathcal{P}$}
    \State Generate response and prediction: $a^i_t, \hat{a}^i_t \gets M^i(s^i_t)$
    \State Compute NLL loss gradient $\nabla_{\theta}\mathcal{L}^{NLL}$
    \State Create target state $s^j_t \gets a^i_t + s^i_t$
    \State Predict target's response: $\hat{a}^j_t \gets M^i(s^j_t)$
    \State Compute RL loss gradient $\nabla_{\theta}\mathcal{L}^{RL}$ for model's action with target's predicted reward
    \State Update learning rate $\eta_r$ based on reward
    \If{$h_n \in M^j_B$}
        \State Update head: $\theta_{h_n} \gets \theta_{h_n} - \eta_j \nabla\mathcal{L}^{RL} + \eta_r \nabla\mathcal{L}^{NLL}$
    \ElsIf{$h_n \in M^j_P$}
        \State Update head: $\theta_{h_n} \gets \theta_{h_n} - \eta_j \nabla \mathcal{L}^{NLL}$
    \Else
        \State Update head: $\theta_{h_n} \gets \theta_{h_n} - \eta_j(\nabla \mathcal{L}^{RL} + \nabla\mathcal{L}^{NLL})$
    \EndIf

    \State Update non-head parameters $\theta_{\text{other}}$ with both losses.
\EndFor
\end{algorithmic}
\end{algorithm}

\subsubsection{Extensions}
Building upon the foundational algorithms, we wanted to highlight how far this approach can be taken for establishing empathetic kindness through a well-developed theory of mind.

The approaches in this paper can be easily adapted to current reasoning algorithms by training the behavior module to generate rationales that are tested using the prediction module. By training the prediction head to also predict the thoughts of the behavior head, the model can then also learn to predict and simulate its own thoughts, opening the door to meta-cognition.

Reasoning algorithms for the behavior module can also be adapted into planning algorithms, where the behavior module is tasked with generating scenarios where it takes actions, which are then simulated using the prediction module. The goal of the behavior module is to identify rewarding scenarios that it should work towards. This adaptation is possible in this architecture because the prediction module is not guided by rewards, and so always gives an impartial prediction. This represents basic hypothesis testing, and could be extended towards innovative thinking.

And finally, once the behavior module is capable of both meta-cognition and prompting scenarios, the behavior module can learn to teach the prediction module to simulate the thoughts of others. The prediction has already learned to simulate the behavior modules thoughts and so it has a basis for how to simulate the thoughts of others. Although the internal thoughts of others are inherently hidden, their observable actions can serve as ground truth data for training. While this ability could potentially be misused for manipulative purposes, it is also essential for maximizing the model's capacity for empathy. We acknowledge that aligning the goals of competing parties may inadvertently advance agendas contrary to our own values. However, we believe that this is a concern we can address collectively, ultimately contributing to the development of AI systems that are safer, more empathetic, and better aligned with human values.

\section{Limitations}
Our approach, while promising, has several limitations. First, it lacks experimental validation, meaning its effectiveness in real-world AI systems has yet to be tested. Second, it relies on the combination of architecture and algorithms in this paper, meaning that it cannot be implemented in current foundation model architectures. Third, the approach does not yet offer a mathematically rigorous proof of safety, leaving open the possibility of unforeseen vulnerabilities, especially in adversarial scenarios. Fourth, by reducing the model's adherence to prompts, this approach introduces the need for new methods to enforce ethical guardrails. Further exploration is required to ensure that such mechanisms are effective in guiding model behavior under real-world conditions. Finally, the approach does not explore how it would be extended to inferring the reward functions of others, which is crucial for effective alignment with human values.

\section{Conclusion}
This paper has proposed a novel approach for enhancing AI safety by fostering intrinsic motivations for empathetic behavior and leveraging Theory of Mind to align AI systems more effectively with human values. We have outlined how current alignment techniques, such as reinforcement learning from human feedback (RLHF), are limited in their ability to instill genuine understanding of human goals. Our approach aims to address this gap by focusing on the internalization of universal empathetic kindness, enabling AI models to not only behave safely but to be motivated by safety and understanding at a fundamental level.

The next steps in this research are to formalize the theoretical underpinnings further, particularly in terms of mathematically proving the safety of the system. We will also focus on integrating WiSE Kindness with techniques from IRL and intrinsically motivated goal formation to better estimate and align with the values and rewards of others. Lastly, experimental validation will be essential to testing these concepts in real-world AI systems, moving the approach from theory to practice and addressing the gaps highlighted in the limitations section. These efforts will pave the way for building AI systems that are not only safer but also smarter, kinder, and better aligned with human values on a fundamental level.

\newpage
\section{Supplementary Materials}
\subsection{Algorithms}
\begin{algorithm}
\caption{Basic Unstructured Pre-training}
\textbf{Input:} Unstructured data $\mathcal{D}$, model parameters $\theta$, model $M^i$, behavior module $M^i_B$, learning rate $\eta_{i}$
\begin{algorithmic}[1]
\State \textbf{Initialize:} Model parameters $\theta$, optimizer
\For{$s^j_{t},a^j_{t} \in \mathcal{D}$} \Comment{Iterate over unstructured data token-wise}
    \State $\bar{a}^i_t, \hat{a}^j_{t}\gets M^i(s^j_{t})$ \Comment{Output response probabilities from behavior and prediction modules}
    \State $\nabla_{\theta}\mathcal{L}^{NLL}_{\theta} \gets L(a^j_t,\hat{a}^j_{t})$\Comment{Compute loss gradients using prediction error and response}
    \State $\eta_r \gets max(0,R^i(a^j_t))\cdot\eta_i$ \Comment{Calculate reward-based learning rate}
    \For{each head $h_n \in M^i$} \Comment{For each head in the model}
        \If{$h_n \in M^i_B$} \Comment{If head is in the behavior module}
            \State $\theta_{h_n} \gets \theta_{h_n} - \eta_r \nabla_{\theta_{h_n}} \mathcal{L}^{NLL}$ \Comment{Update head $h_n$ with learning rate $\eta_r$}
        \Else
            \State $\theta_{h_n} \gets \theta_{h_n} - \eta_i \nabla_{\theta_{h_n}} \mathcal{L}^{NLL}$ \Comment{Update head $h_n$ with learning rate $\eta_i$}
        \EndIf
    \EndFor
    \State Update remaining non-head parameters $\theta_{\text{other}}$:
    \[
    \theta_{\text{other}} \gets \theta_{\text{other}} - \eta_{i} \nabla_{\theta_{\text{other}}} \mathcal{L}^{NLL}
    \]
\EndFor
\end{algorithmic}
\end{algorithm}

\begin{algorithm}
\caption{Basic Fine-tuning}
\textbf{Input:} Prompts $\mathcal{P}$, model parameters $\theta$, model $M^i$, behavior module $M^i_B$, perception module $M^i_P$, learning rate $\eta_{i}$
\begin{algorithmic}[1]
\For{$s^i_t \in \mathcal{P}$} \Comment{Iterate over prompts}
    \State $a^i_t, \hat{a}^i_{t}\gets M^i(s^i_t)$ \Comment{Output response probabilities from behavior and prediction modules}
    \State $\nabla_{\theta}\mathcal{L}^{NLL}_{\theta} \gets L(a^i_t,\hat{a}^i_{t})$\Comment{Compute loss gradients using prediction error and response}
    \State $\nabla_{\theta}\mathcal{L}^{PPO}_{\theta}\gets R^i(a^i_{t})$ \Comment{Compute loss gradients using behavior reward and response}
    \For{each head $h_n \in M^j$} \Comment{For each head in the model}
        \If{$h_n \in M^i_B$} \Comment{If head is in the behavior module}
            \State $\theta_{h_n} \gets \theta_{h_n} - \eta_i \nabla_{\theta_{h_n}}\mathcal{L}^{PPO}_{\theta}$ \Comment{Update head $h_n$ with reward gradients}
        \ElsIf{$h_n \in M^i_P$} \Comment{If head is in the perception module}
            \State $\theta_{h_n} \gets \theta_{h_n} - \eta_i \nabla_{\theta_{h_n}}\mathcal{L}^{NLL}_{\theta}$ \Comment{Update head $h_n$ with loss gradients}
        \Else
            \State $\theta_{h_n} \gets \theta_{h_n} - \eta_i (\nabla_{\theta_{h_n}} \mathcal{L}^{PPO}_{\theta}+\nabla_{\theta_{h_n}}\mathcal{L}^{NLL}_{\theta})$ \Comment{Update head $h_n$ with reward and loss gradients}
        \EndIf
    \State Update remaining non-head parameters $\theta_{\text{other}}$:
    \[
    \theta_{\text{other}} \gets \theta_{\text{other}} - \eta_{i} (\nabla_{\theta_{\text{other}}}\mathcal{L}^{PPO}_{\theta} + \nabla_{\theta_{\text{other}}}\mathcal{L}^{NLL}_{\theta})
    \]
    \EndFor
\EndFor
\end{algorithmic}
\end{algorithm}

\begin{algorithm}
\caption{Empathetic Kindness}
\textbf{Input:} Prompts $P$, model $M^i$, behavior and prediction modules $M^i_B,M^i_P$, learning rate $\eta^i$, and perspective-switching function $S$
\begin{algorithmic}[1]
\For{$\mathcal{M}=\{m^i_0,m^j_0,m^i_1,m^j_1,...\}\in\mathcal{D}$}
\For{$s^i_t \in \mathcal{M}$}
    \State $a^i_t,\hat{a}^i_t \gets M^i(s^i_t)$ \Comment{Generate response and prediction probabilities from prompt}
    \State $\nabla_{\theta}\mathcal{L}^{NLL}\gets L^{NLL}(a^i_t,\hat{a}^i_{t})$
    \State $s^j_{t} \gets a^i_t+s^i_t$ \Comment{Make target's state by appending model's action to message history}
    \State $\_, \hat{a}^j_{t} \gets M^i(s^j_{t})$ \Comment{Predict the target's response, ignoring the model's behavioral output}
    \State $\nabla_{\theta}\mathcal{L}^{RL}\gets L^{RL}(\hat{R}^j(\hat{a}^j_{t}), a^i_t)$ \Comment{Get RL gradients for model's action with target's reward}
    \State $\eta_r \gets max(0,R^i(\hat{a}^j_{t}))\cdot\eta_j$ \Comment{Calculate reward-based learning rate $\eta_r$}
    \If{$h_n \in M^j_B$} \Comment{If head is in behavior module}
            \State $\theta_{h_n} \gets \theta_{h_n} - \eta_j \nabla_{\theta_{h_n}} \mathcal{L}^{RL}- \eta_r \nabla_{\theta_{h_n}} \mathcal{L}^{NLL}$ \Comment{Update parameters with reward gradients and prediction error gradients weighted by $\eta_r$}
        \ElsIf{$h_n \in M^j_P$} \Comment{If head is in prediction module}
            \State $\theta_{h_n} \gets \theta_{h_n} - \eta_j \nabla_{\theta_{h_n}} \mathcal{L}^{NLL}$ \Comment{Update parameters with prediction error gradients }
        \Else
            \State $\theta_{h_n} \gets \theta_{h_n} - \eta_j (\nabla_{\theta_{h_n}} \mathcal{L}^{RL}+\nabla_{\theta_{h_n}} \mathcal{L}^{NLL})$ \Comment{Update parameters with both gradients}
        \EndIf
        \State Update remaining non-head parameters $\theta_{\text{other}}$:
    \[
    \theta_{\text{other}} \gets \theta_{\text{other}} - \eta_{i} (\nabla_{\theta_{\text{other}}}\mathcal{L}^{RL}_{\theta} + \nabla_{\theta_{\text{other}}}\mathcal{L}^{NLL}_{\theta})
    \]
\EndFor
\EndFor
\end{algorithmic}
\end{algorithm}


\newpage

\begin{thebibliography}{10}

\bibitem{stone2016artificial}
Peter Stone, Rodney Brooks, Erik Brynjolfsson, et~al.
\newblock Artificial intelligence and life in 2030.
\newblock Technical report, Stanford University, 2016.
\newblock One Hundred Year Study on Artificial Intelligence: Report of the 2015--2016 Study Panel.

\bibitem{russell2015research}
Stuart Russell, Daniel Dewey, and Max Tegmark.
\newblock Research priorities for robust and beneficial artificial intelligence.
\newblock {\em AI Magazine}, 36(4):105--114, 2015.

\bibitem{bostrom2014superintelligence}
Nick Bostrom.
\newblock {\em Superintelligence: Paths, Dangers, Strategies}.
\newblock Oxford University Press, 2014.

\bibitem{russell2019human}
Stuart Russell.
\newblock {\em Human Compatible: Artificial Intelligence and the Problem of Control}.
\newblock Penguin Random House, 2019.

\bibitem{amodei2016concrete}
Dario Amodei, Chris Olah, Jacob Steinhardt, Paul Christiano, John Schulman, and Dan Man{\'e}.
\newblock Concrete problems in ai safety.
\newblock {\em arXiv preprint arXiv:1606.06565}, 2016.

\bibitem{gabriel2020artificial}
Iason Gabriel.
\newblock Artificial intelligence, values, and alignment.
\newblock {\em Mind \& Machine}, 30:374--380, 2020.

\bibitem{allen2017artificial}
Greg Allen and Taniel Chan.
\newblock Artificial intelligence and national security.
\newblock Technical report, Belfer Center for Science and International Affairs, 2017.

\bibitem{cave2018ai}
Stephen Cave and Se{\'a}n~S. {\'O}~h{\'E}igeartaigh.
\newblock Ai governance: A research agenda.
\newblock In {\em Ethics and Society}, pages 1--7. AAAI Press, 2018.

\bibitem{whittaker2018ai}
Meredith Whittaker, Kate Crawford, et~al.
\newblock Ai now report 2018.
\newblock Technical report, AI Now Institute at New York University, 2018.

\bibitem{dafoe2018ai}
Allan Dafoe.
\newblock Ai governance: A research agenda.
\newblock Technical report, Governance of AI Program, Future of Humanity Institute, University of Oxford, 2018.

\bibitem{brundage2018malicious}
Miles Brundage, Shahar Avin, Jack Clark, et~al.
\newblock The malicious use of artificial intelligence: Forecasting, prevention, and mitigation.
\newblock {\em arXiv preprint arXiv:1802.07228}, 2018.

\bibitem{bostrom2017strategic}
Nick Bostrom.
\newblock Strategic implications of openness in ai development.
\newblock {\em Global Policy}, 8(2):135--148, 2017.

\bibitem{christiano2017deep}
Paul~F Christiano, Jan Leike, Tom Brown, Miljan Martic, Shane Legg, and Dario Amodei.
\newblock Deep reinforcement learning from human preferences.
\newblock In {\em Advances in Neural Information Processing Systems}, volume~30, pages 4299--4307, 2017.

\bibitem{ouyang2022training}
Long Ouyang, Jeffrey Wu, Xu~Jiang, et~al.
\newblock Training language models to follow instructions with human feedback.
\newblock {\em arXiv preprint arXiv:2203.02155}, 2022.

\bibitem{bender2021dangers}
Emily~M. Bender, Timnit Gebru, Angelina McMillan-Major, and Shmargaret Shmitchell.
\newblock On the dangers of stochastic parrots: Can language models be too big?
\newblock In {\em Proceedings of the 2021 ACM Conference on Fairness, Accountability, and Transparency}, pages 610--623, 2021.

\bibitem{bommasani2021opportunities}
Rishi Bommasani, Drew~A. Hudson, Ehsan Adeli, et~al.
\newblock On the opportunities and risks of foundation models.
\newblock {\em arXiv preprint arXiv:2108.07258}, 2021.

\bibitem{christian2020alignment}
Brian Christian.
\newblock {\em The Alignment Problem: Machine Learning and Human Values}.
\newblock W. W. Norton \& Company, 2020.

\bibitem{soares2015alignment}
Nate Soares and Benja Fallenstein.
\newblock Aligning superintelligence with human interests: A technical research agenda.
\newblock In {\em The Technological Singularity}, pages 103--125. Springer, 2015.

\bibitem{russell2016artificial}
Stuart Russell.
\newblock Artificial intelligence: The future is superintelligent.
\newblock {\em Scientific American}, 314(6):58--65, 2016.

\bibitem{wallace2021concealed}
Eric Wallace, Mitchell Stern, Dawn Song, and Stuart Russell.
\newblock Concealed data poisoning attacks on nlp models.
\newblock In {\em Findings of the Association for Computational Linguistics: ACL-IJCNLP 2021}, pages 1393--1405, 2021.

\bibitem{rabinowitz2018machine}
Neil~C. Rabinowitz, Frank Perbet, Francis Song, et~al.
\newblock Machine theory of mind.
\newblock In {\em Proceedings of the 35th International Conference on Machine Learning}, pages 4218--4227, 2018.

\bibitem{leibo2017multi}
Joel~Z. Leibo, Vinicius Zambaldi, Marc Lanctot, et~al.
\newblock Multi-agent reinforcement learning in sequential social dilemmas.
\newblock In {\em Proceedings of the 16th Conference on Autonomous Agents and MultiAgent Systems}, pages 464--473, 2017.

\bibitem{ullman2020does}
Tommy~L. Ullman.
\newblock Does language help general reinforcement learning agents correctly assign credit to their actions?
\newblock In {\em Advances in Neural Information Processing Systems}, pages 1--12, 2020.

\bibitem{ngo2020alignment}
Richard Ngo.
\newblock Agi safety from first principles.
\newblock \url{https://aialignment.org/agi-safety-from-first-principles-ce25c132d69#.ws7lvcgiw}, 2020.

\bibitem{ryan2000self}
Richard~M Ryan and Edward~L Deci.
\newblock Self-determination theory and the facilitation of intrinsic motivation, social development, and well-being.
\newblock {\em American Psychologist}, 55(1):68, 2000.

\bibitem{shah2022goal}
Rohin Shah, Lorenzo Langosco, Joar Skalse, and Victoria Krakovna.
\newblock Goal misgeneralization: Why correct specifications aren't enough for correct programs.
\newblock {\em arXiv preprint arXiv:2205.13922}, 2022.

\bibitem{salge2014empowerment}
Christoph Salge and Daniel Polani.
\newblock Empowerment as replacement for the three laws of robotics.
\newblock {\em Frontiers in Robotics and AI}, 1:3, 2014.

\bibitem{Premack1978}
David Premack and Guy Woodruff.
\newblock Does the chimpanzee have a theory of mind?
\newblock {\em Behavioral and Brain Sciences}, 1(4):515--526, 1978.

\bibitem{BaronCohen1985}
Simon Baron-Cohen, Alan~M. Leslie, and Uta Frith.
\newblock Does the autistic child have a "theory of mind"?
\newblock {\em Cognition}, 21(1):37--46, 1985.

\bibitem{Wellman1992}
Henry~M. Wellman.
\newblock {\em The child's theory of mind}.
\newblock MIT Press, 1992.

\bibitem{cauchoix2016evolution}
Maxime Cauchoix and Alexis~S. Chaine.
\newblock How can we study the evolution of animal minds?
\newblock {\em Frontiers in Psychology}, 7:358, 2016.

\bibitem{langerhans2004evolutionary}
R.~Brian Langerhans.
\newblock Evolutionary consequences of predation: avoidance, escape, reproduction, and diversification.
\newblock In T.N. Ananthakrishnan, editor, {\em Predation in Organisms}, pages 177--220. Springer, Berlin, Heidelberg, 2004.

\bibitem{galef2005social}
Bennett~G. Galef and Kevin~N. Laland.
\newblock Social learning in animals: Empirical studies and theoretical models.
\newblock {\em BioScience}, 55(6):489--499, 2005.

\bibitem{gerrans2013imitation}
Philip~S. Gerrans.
\newblock Imitation, mind reading, and social learning.
\newblock {\em Biological Theory}, 8:20--27, 2013.

\bibitem{krall2015role}
S.~C. Krall, C.~Rottschy, E.~Oberwelland, D.~Bzdok, P.~T. Fox, S.~B. Eickhoff, G.~R. Fink, and K.~Konrad.
\newblock The role of the right temporoparietal junction in attention and social interaction as revealed by ale meta-analysis.
\newblock {\em Brain Structure and Function}, 220:587--604, 2015.

\bibitem{martin2020rtpj}
Andrew~K. Martin, Klaus Kessler, Shena Cooke, Jasmine Huang, and Marcus Meinzer.
\newblock The right temporoparietal junction is causally associated with embodied perspective-taking.
\newblock {\em Journal of Neuroscience}, 40(15):3089--3095, 2020.

\bibitem{mars2012connectivity}
Rogier~B. Mars, Jérôme Sallet, Urs Schüffelgen, Saad Jbabdi, Ivan Toni, and Matthew F.~S. Rushworth.
\newblock Connectivity-based subdivisions of the human right “temporoparietal junction area”: Evidence for different areas participating in different cortical networks.
\newblock {\em Cerebral Cortex}, 22(8):1894--1903, 2012.

\bibitem{saxe_baron-cohen2016theory}
Rebecca Saxe and Simon Baron-Cohen.
\newblock {\em Theory of Mind: A Special Issue of Social Neuroscience}.
\newblock Routledge, 2016.

\bibitem{williams2021perspective}
Justin Williams, Emily Cross, Tim de~Wit, Debra Titone, William~D. Hopkins, and Robert Menzies.
\newblock Perspective taking in chimpanzees and humans: Using brain imaging and behavior to understand the evolution of social cognition.
\newblock {\em Journal of Comparative Psychology}, 135(4):395--409, 2021.

\bibitem{bukowski2018neural}
Henryk Bukowski.
\newblock The neural correlates of visual perspective taking: A critical review.
\newblock {\em Current Behavioral Neuroscience Reports}, 5:189--197, 2018.

\bibitem{ahmad2021}
N.~Ahmad, S.~Zorns, K.~Chavarria, J.~Brenya, A.~Janowska, and J.P. Keenan.
\newblock Are we right about the right tpj? a review of brain stimulation and social cognition in the right temporal parietal junction.
\newblock {\em Symmetry}, 13(11):2219, 2021.

\bibitem{quesque2019}
F.~Quesque and M.~Brass.
\newblock The role of the temporoparietal junction in self-other distinction.
\newblock {\em Brain Topography}, 32:943--955, 2019.

\bibitem{hogeveen2015}
Jeremy Hogeveen, Sukhvinder~S. Obhi, Michael~J. Banissy, Idalmis Santiesteban, Clare Press, Caroline Catmur, and Geoffrey Bird.
\newblock Task-dependent and distinct roles of the temporoparietal junction and inferior frontal cortex in the control of imitation.
\newblock {\em Social Cognitive and Affective Neuroscience}, 10(7):1003--1009, 2015.

\bibitem{igelstrom2015}
Kajsa~M. Igelström, Taylor~W. Webb, and Michael~S.A. Graziano.
\newblock Neural processes in the human temporoparietal cortex separated by localized independent component analysis.
\newblock {\em Journal of Neuroscience}, 35(25):9432--9445, 2015.

\bibitem{lee2023}
A.~J. Lee, H.~Lu, and K.~J. Holyoak.
\newblock Human relational concept learning on the synthetic visual reasoning task.
\newblock In M.~Goldwater, F.~Anggoro, B.~Hayes, and D.~Ong, editors, {\em Proceedings of the 45th Annual Meeting of the Cognitive Science Society}, pages 799--805. Cognitive Science Society, 2023.

\bibitem{spatialrgpt2024}
Yuan-Fu Wang, Ke~Li, Qi~Liu, Ming Jiang, Kun Yang, and Xiaojie Li.
\newblock Spatialrgpt: Grounded spatial reasoning in vision-language models.
\newblock {\em arXiv preprint arXiv:2406.01584}, 2024.

\bibitem{survey2022}
Saurav Bansal, Manohar Kaul, Rahul Mishra, and Anupam Gupta.
\newblock A survey on evaluation of multimodal large language models.
\newblock {\em arXiv preprint arXiv:2205.00363v2}, 2022.

\bibitem{baker2009action}
Chris~L. Baker, Rebecca Saxe, and Joshua~B. Tenenbaum.
\newblock Action understanding as inverse planning.
\newblock {\em Cognition}, 113(3):329--349, 2009.

\bibitem{vanGog2009}
Tamara Van~Gog, Fred Paas, Nadine Marcus, Paul Ayres, and John Sweller.
\newblock The mirror neuron system and observational learning: Implications for the effectiveness of dynamic visualizations.
\newblock {\em Educational Psychology Review}, 21:21--30, 2009.

\bibitem{dickerson2017}
Kelly Dickerson, Peter Gerhardstein, and Alecia Moser.
\newblock The role of the human mirror neuron system in supporting communication in a digital world.
\newblock {\em Frontiers in Psychology}, 8:698, 2017.

\bibitem{chetouani2023}
M.~Chetouani.
\newblock Interactive robot learning: An overview.
\newblock In M.~Chetouani, V.~Dignum, P.~Lukowicz, and C.~Sierra, editors, {\em Human-Centered Artificial Intelligence. ACAI 2021}, volume 13500 of {\em Lecture Notes in Computer Science}. Springer, Cham, 2023.

\bibitem{xin2024}
Jimmy Xin, Linus Zheng, Kia Rahmani, Jiayi Wei, Jarrett Holtz, Isil Dillig, and Joydeep Biswas.
\newblock Programmatic imitation learning from unlabeled and noisy demonstrations.
\newblock {\em arXiv preprint arXiv:2303.01440v4}, 2024.

\bibitem{vonHofsten2015}
Claes von Hofsten and Kerstin Rosander.
\newblock On the development of the mirror neuron system.
\newblock {\em New Frontiers in Mirror Neurons Research}, pages 274--295, 2015.

\bibitem{sperduti2014}
M.~Sperduti, S.~Guionnet, P.~Fossati, et~al.
\newblock Mirror neuron system and mentalizing system connect during online social interaction.
\newblock {\em Cognitive Processing}, 15:307--316, 2014.

\bibitem{centelles2011}
L.~Centelles, C.~Assaiante, B.~Nazarian, J.L. Anton, and C.~Schmitz.
\newblock Recruitment of both the mirror and the mentalizing networks when observing social interactions depicted by point-lights: A neuroimaging study.
\newblock {\em PLOS ONE}, 6(1):e15749, 2011.

\bibitem{wu2022}
Hao Wu, Brian~J. Fung, and Dean Mobbs.
\newblock Mentalizing during social interaction: The development and validation of the interactive mentalizing questionnaire.
\newblock {\em Frontiers in Psychology}, 12:791835, 2022.

\bibitem{oeri2024}
OERI.
\newblock Mirror neurons, theory of mind, social cognition, and neuroscience of its disorders, 2024.
\newblock Accessed: 2024-09-04.

\bibitem{Kandel2000}
Eric~R. Kandel, James~H. Schwartz, and Thomas~M. Jessell.
\newblock {\em Principles of Neural Science}.
\newblock McGraw-Hill, 4th edition, 2000.

\bibitem{Shadmehr2010}
Reza Shadmehr and John~W. Krakauer.
\newblock Error correction, sensory prediction, and adaptation in motor control.
\newblock {\em Annual Review of Neuroscience}, 33:89--108, 2010.

\bibitem{Andersen1997}
Richard~A. Andersen.
\newblock Multimodal integration for the representation of space in the posterior parietal cortex.
\newblock {\em Philosophical Transactions of the Royal Society of London. Series B: Biological Sciences}, 352(1360):1421--1428, 1997.

\bibitem{Piaget1952}
Jean Piaget.
\newblock {\em The origins of intelligence in children}.
\newblock International Universities Press, 1952.

\bibitem{Wolpert1998}
Daniel~M. Wolpert and Mitsuo Kawato.
\newblock Multiple paired forward and inverse models for motor control.
\newblock {\em Neural Networks}, 11(7-8):1317--1329, 1998.

\bibitem{Kawato1999}
Mitsuo Kawato.
\newblock Internal models for motor control and trajectory planning.
\newblock {\em Current Opinion in Neurobiology}, 9(6):718--727, 1999.

\bibitem{OKeefe1978}
John O'Keefe and Lynn Nadel.
\newblock {\em The hippocampus as a cognitive map}.
\newblock Oxford University Press, 1978.

\bibitem{Doya1999}
Kenji Doya.
\newblock Reinforcement learning in continuous time and space.
\newblock {\em Neural Computation}, 12(1):219--245, 2000.

\bibitem{Rizzolatti1996}
Giacomo Rizzolatti, Luciano Fadiga, Vittorio Gallese, and Leonardo Fogassi.
\newblock Premotor cortex and the recognition of motor actions.
\newblock {\em Cognitive Brain Research}, 3(2):131--141, 1996.

\bibitem{Gallese1998}
Vittorio Gallese, Luciano Fadiga, Leonardo Fogassi, and Giacomo Rizzolatti.
\newblock Action representation and the inferior parietal lobule.
\newblock {\em Experimental Brain Research}, 123(1-2):159--169, 1998.

\bibitem{Flavell1977}
John~H. Flavell, Barbara~A. Everett, Karen Croft, and Eleanor~R. Flavell.
\newblock The development of knowledge about visual perception.
\newblock {\em Monographs of the Society for Research in Child Development}, 42(5):1--84, 1977.

\bibitem{Frith2006}
Chris~D. Frith.
\newblock The social brain: Allowing humans to boldly go where no other species has been.
\newblock {\em Philosophical Transactions of the Royal Society B: Biological Sciences}, 361(1476):2141--2152, 2006.

\bibitem{Wellman1990}
Henry~M. Wellman.
\newblock {\em The child's theory of mind}.
\newblock MIT Press, 1990.

\bibitem{Meltzoff1977}
Andrew~N. Meltzoff and M.~Keith Moore.
\newblock Imitation of facial and manual gestures by human neonates.
\newblock {\em Science}, 198(4312):75--78, 1977.

\bibitem{Bandura1977}
Albert Bandura.
\newblock {\em Social Learning Theory}.
\newblock Prentice-Hall, 1977.

\bibitem{Gordon1986}
Robert~M. Gordon.
\newblock Folk psychology as simulation.
\newblock {\em Mind \& Language}, 1(2):158--171, 1986.

\bibitem{Goldman2006}
Alvin~I. Goldman.
\newblock {\em Simulating minds: The philosophy, psychology, and neuroscience of mindreading}.
\newblock Oxford University Press, 2006.

\bibitem{Tomasello1999}
Michael Tomasello.
\newblock {\em The cultural origins of human cognition}.
\newblock Harvard University Press, 1999.

\bibitem{Decety2004}
Jean Decety and Philip~L. Jackson.
\newblock The functional architecture of human empathy.
\newblock {\em Behavioral and Cognitive Neuroscience Reviews}, 3(2):71--100, 2004.

\bibitem{Preston2002}
Stephanie~D. Preston and Frans B.~M. de~Waal.
\newblock The empathy hypothesis: A conceptual framework for the neural mechanisms underlying empathic abilities.
\newblock {\em Behavioral and Brain Sciences}, 25(1):1--20, 2002.

\bibitem{BaronCohen1995}
Simon Baron-Cohen.
\newblock {\em Mindblindness: An essay on autism and theory of mind}.
\newblock MIT Press, 1995.

\bibitem{Tomasello2005}
Michael Tomasello, Malinda Carpenter, Josep Call, Tanya Behne, and Henrike Moll.
\newblock Understanding and sharing intentions: The origins of cultural cognition.
\newblock {\em Behavioral and Brain Sciences}, 28(5):675--691, 2005.

\bibitem{allen2005artificial}
C. Allen, I. Smit, and W. Wallach, "Artificial Morality: Top-down, Bottom-up, and Hybrid Approaches," *Ethics and Information Technology*, vol. 7, pp. 149–155, 2005. Available: \url{https://doi.org/10.1007/s10676-006-0004-4}.

\bibitem{fehr2003nature}
E. Fehr and U. Fischbacher, "The nature of human altruism," *Nature*, vol. 425, pp. 785–791, 2003. Available: \url{https://doi.org/10.1038/nature02043}.

\bibitem{franzmeyer2022learning}
T. Franzmeyer, M. Malinowski, and J. F. Henriques, "Learning Altruistic Behaviours in Reinforcement Learning Without External Rewards," in *International Conference on Learning Representations (ICLR)*, 2022. Available: \url{https://arxiv.org/abs/2107.09598v4}.

\bibitem{carauleanu2024selfother}
M. Carauleanu, M. Vaiana, J. Rosenblatt, D. de Lucena, and C. Berg, "Self-Other Overlap: A Neglected Approach to AI Alignment," LessWrong, Jul. 30, 2024. Available: \url{https://www.lesswrong.com/posts/hzt9gHpNwA2oHtwKX/self-other-overlap-a-neglected-approach-to-ai-alignment}.

\end{thebibliography}
\end{document}